\documentclass[10pt,twocolumn,letterpaper]{article}

\usepackage{wacv}
\usepackage{times}
\usepackage{epsfig}
\usepackage{graphicx}
\usepackage{amsmath}
\usepackage{amssymb}
\usepackage{booktabs}
\usepackage{graphicx, subfigure}
\usepackage{paralist}
\usepackage{pifont}
\newcommand{\cmark}{\ding{51}}%
\newcommand{\xmark}{\ding{55}}%

\usepackage[breaklinks=true,bookmarks=false]{hyperref}

\begin{document}

\title{Group Activity Recognition in Basketball Tracking Data - \\ Neural Embeddings in Team Sports (NETS)}

\author{Sandro Hauri\\
Temple University\\
{\tt\small sandro.hauri@temple.edu}
\and
Slobodan Vucetic\\
Temple University\\
{\tt\small vucetic@temple.edu}
}

\maketitle
\thispagestyle{empty}

\begin{abstract}
    \noindent Like many team sports, basketball involves two groups of players who engage in collaborative and adversarial activities to win a game. Players and teams are executing various complex strategies to gain an advantage over their opponents. Defining, identifying, and analyzing different types of activities is an important task in sports analytics, as it can lead to better strategies and decisions by the players and coaching staff. The objective of this paper is to automatically recognize basketball group activities from tracking data representing locations of players and the ball during a game. We propose a novel deep learning approach for group activity recognition (GAR) in team sports called NETS. To efficiently model the player relations in team sports, we combined a Transformer-based architecture with LSTM embedding, and a team-wise pooling layer to recognize the group activity. Training such a neural network generally requires a large amount of annotated data, which incurs high labeling cost. To address scarcity of manual labels, we generate weak-labels and pretrain the neural network on a self-supervised trajectory prediction task. We used a large tracking data set from 632 NBA games to evaluate our approach. The results show that NETS is capable of learning group activities with high accuracy, and that self- and weak-supervised training in NETS have a positive impact on GAR accuracy.
\end{abstract}

\section{Introduction}

\noindent With the recent advances in sensing technology, there is an unprecedented amount of tracking data available for sports analytics, such as Hawk-eye in tennis \cite{owens2003hawk}, various tracking systems in soccer (ChyronHego, Stats LLC, SciSports), and SPORTLOGiQ in ice hockey. The National Basketball Association (NBA) mandated the installation of a tracking system based on computer vision called SportVU \cite{stats2019stats} in their sports arenas to collect data about the movements of players and the ball at 25Hz. This data is shared with all NBA teams to ensure equity, with an implicit understanding that the teams with the best ability to benefit from this data will not only improve their own competitiveness, but will advance innovation in the way basketball is played.

\begin{figure}[t]
\centering
\subfigure[Start of extracted play sequence. The ball handler $a_1$ will dribble to the right and attack.]{
\label{fig:p&r_a}\includegraphics[width=0.45\linewidth]{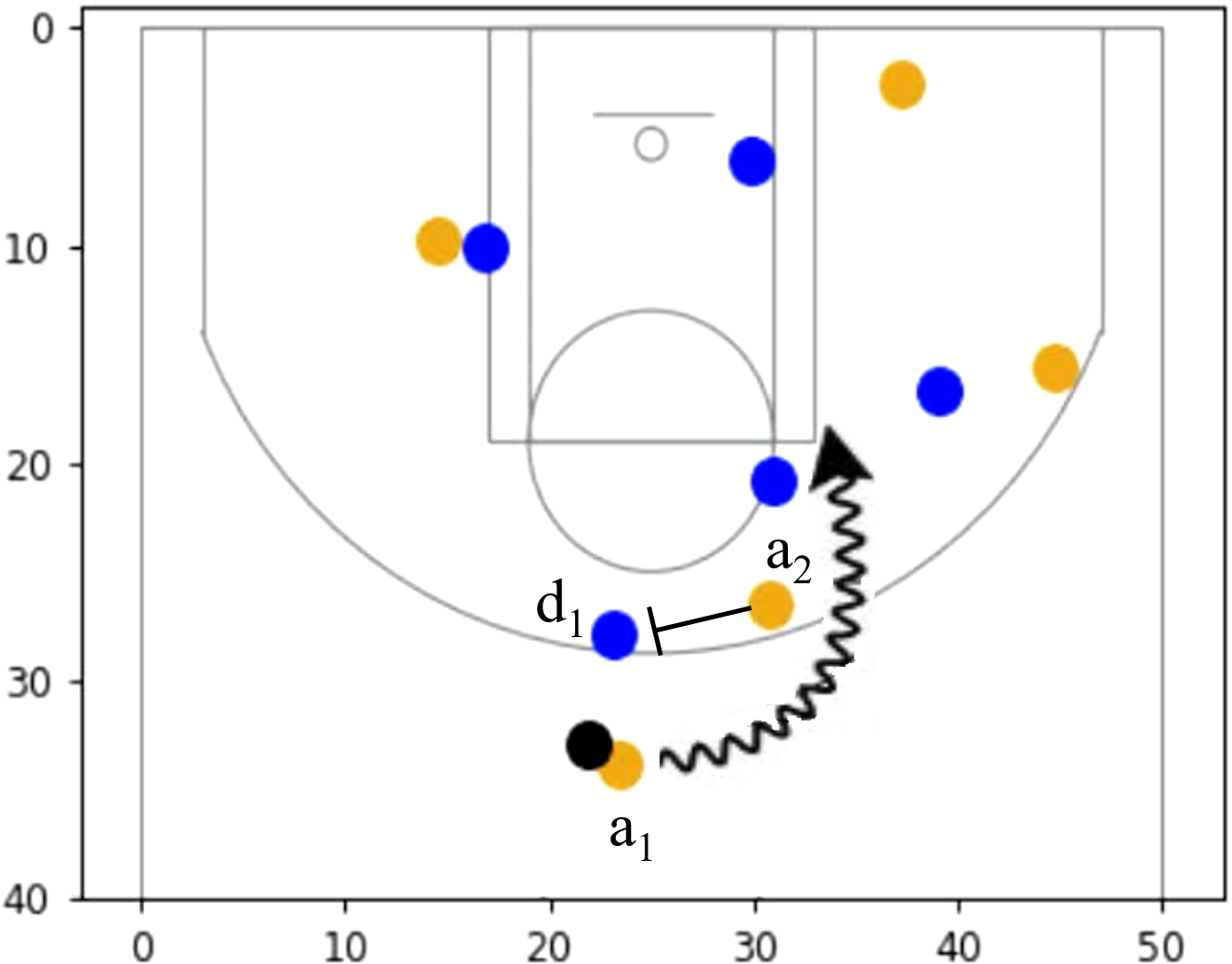}
}\hfill
\subfigure[The attacker $a_2$ blocks the path of the defender. The ball handler $a_1$ keeps possession of the ball.]{
\label{fig:p&r_b}\includegraphics[width=0.45\linewidth]{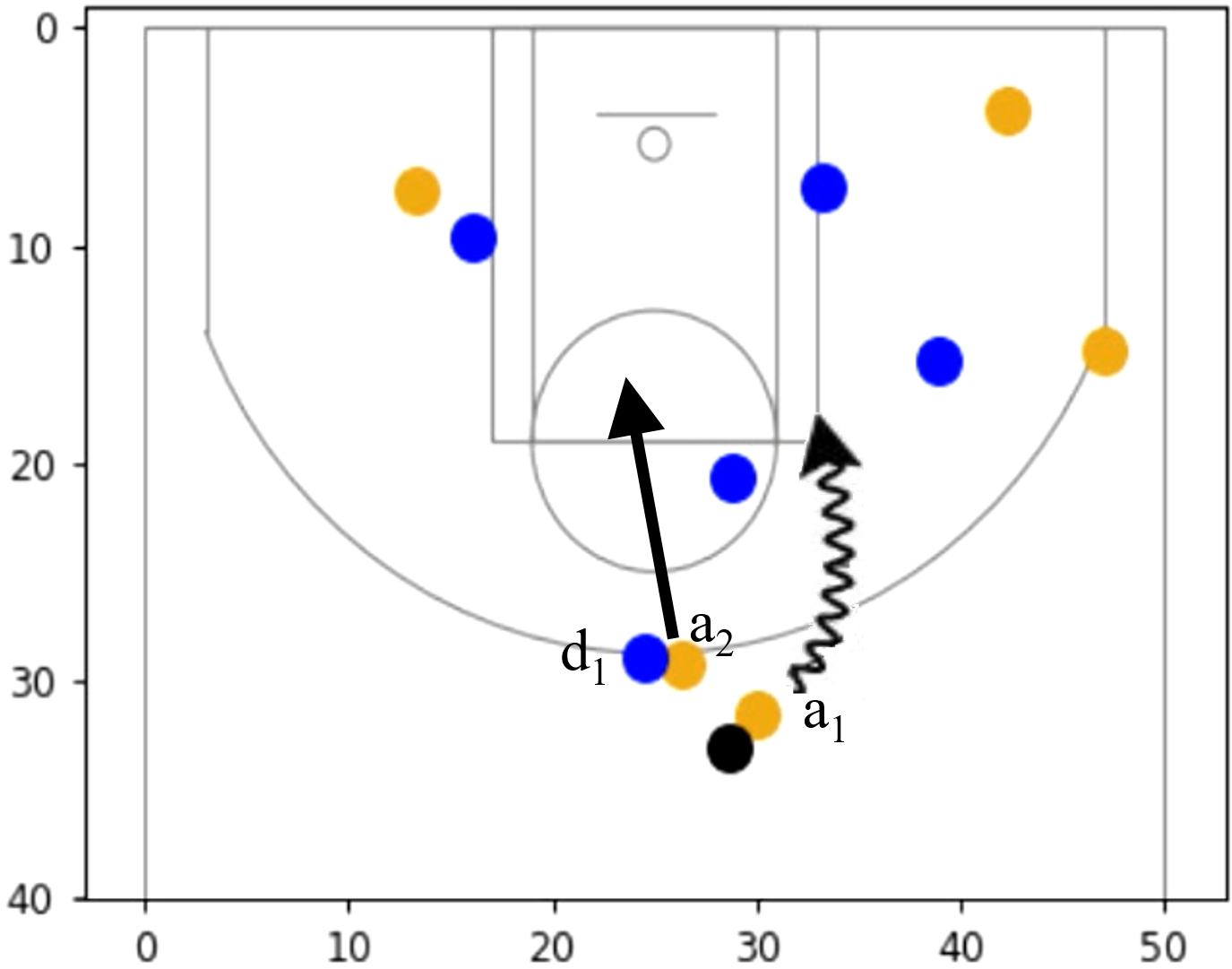}
}
\caption{A classic pick-and-roll. Attackers are displayed in gold, defenders are shown in blue and the ball in black. The roll man $a_2$ helps the ball handler $a_1$ by blocking the defender $d_1$.}
\vspace{-3mm}
\label{fig:pick-and-roll}
\end{figure}

Single player actions in basketball, such as shot-taking, can be recognized with rules programmed by an expert. In contrast, group activities involve multiple players that perform a coordinated action dependent on their combined movements over time, which is much more challenging to describe with hand-written rules. As an example, let us consider a commonly observed tactic in basketball called pick-and-roll. For this tactic, two offensive players are working together to block the best path of a defender (see Figure \ref{fig:pick-and-roll}). To identify this action, we have to understand key concepts of basketball, such as ball possession, defensive assignments, and cooperation of offensive players. To pose a further challenge, there is a closely related tactic called handoff, where two offensive players cross paths and transfer the ball possession with a short pass. Separating the two types of activities can be a challenge, because the setup of a pick-and-roll and a handoff is very similar.

Manually labeling play sequences is time consuming, because it requires watching all the games with a lot of focus while paying attention to multiple players simultaneously. There are also edge cases that some experts would include and others would not. As a consequence, existing manually labeled datasets of pick-and-rolls only include roughly 1,000 play sequences \cite{nunes2016influence, koutsouridis2018efficacy}, which is not enough data to train powerful neural network (NN) models. One option to supply more labels is to write rule-based code to recognize actions, yielding weak-labels. As we will elaborate in Section 4.2 and show in the results in Section 6.5, this is not a simple process and even after a lot of effort the quality of hand-written code is inferior to manual labeling. Since rule-based weak-labeling can quickly provide labels for an entire data set, our hypothesis is that those weak-labels can be exploited during training of a deep learning model. Additionally, we propose to use self-supervised pretraining with trajectory predictions. Our hypothesis is that an NN trained to predict trajectories inherently learns about the behavior of players and groups of players. Thus, we expect that an NN pretrained in this way could be successfully fine-tuned for Group Activity Recognition (GAR). In our approach, we combine pretraining on trajectory prediction task, fine-tuning using weak labels and further fine-tuning using manual labels sequentially.

We propose the use of a transformer architecture \cite{vaswani2017attention} that creates context aware embeddings for each player, which are gathered through a team-wise pooling layer and call our model Neural Embeddings for Team Sports (NETS). Our model uses Long Short-Term Memory (LSTM) \cite{hochreiter1997long} to embed the time series to allow the modeling of the temporal aspect of the data. The team-wise pooling layer allows to create play embeddings that are permutation-invariant to the input order of players within a team.

Our main contributions are:
\begin{compactitem}
    \item A multistage training approach to pretrain on self-supervised data, with subsequent fine-tuning on weak-labels and manual labels to learn complex GAR.
    \item A novel Transformer-based NN architecture to create predictions and embeddings that are invariant to player permutation within a team.
    \item Extensive experimental evaluation on large basketball tracking data.
\end{compactitem}

\section{Related Work}
\label{RelatedWork}

\noindent GAR has gained significant interest from computer vision research with practical applications such as social role understanding, security surveillance, and sports analysis \cite{wu2021comprehensive}. While some researchers study GAR based on tracking information in pedestrian data \cite{zhou2011recognizing}, surveillance video \cite{ni2009recognizing}, or exercising activity \cite{jaouedi2020new}, not many works in sports analytics rely only on tracking data \cite{sicilia2019deephoops}. Group activities have been studied in volleyball with the team actions spike, set, pass, and winpoint with a popular benchmark dataset based on short video segments \cite{ibrahim2016hierarchical}. Various deep learning approaches improved classification accuracy on this dataset, using state of the art image recognition tools to solve this task \cite{yuan2021spatio, xu2020group, duan2021revisiting, zappardino2021learning}. GAR was used in soccer analytics to recognize pass, reception, and shot in a given play sequence \cite{Sanford_2020_CVPR_Workshops}, where the authors use a combination of tracking data and the original video to solve the problem. We believe that using only tracking data in team sports can be advantageous because it reduces the amount of data significantly,  making it possible to train on much larger data. Group activities in sports happen in a standardized environment, where the geometry of the court can be learned without the need for the image context.

Human trajectory prediction is an active field of research with many applications, such as navigation of autonomous vehicles, planning of transportation systems, traffic operations, and many more \cite{kothari2020human}. Modeling the implicit social interactions from trajectory data is a challenging task with a long history of research from social force models \cite{helbing1995social}, locally optimal collision avoidance \cite{van2008reciprocal}, to the current state of the art of modeling human interactions using deep neural networks \cite{gupta2018social, ivanovic2019trajectron, kosaraju2019social, vemula2018social}. Trajectory prediction on NBA data has found much interest in the computer vision community, with notable examples using Variational AutoEncoders (VAE) to model the players' movements \cite{sun2019stochastic, zhan2018generating, zheng2016generating}. However, VAEs are notoriously difficult to train \cite{dai2020usual}, and it can be difficult to tune the parameters to generate realistic samples \cite{fu2019cyclical}. Another deep learning method to predict NBA trajectories uses an LSTM-base approach \cite{hauri2020multi}.

We propose the use of self-supervised learning to help training GAR models. Self-supervised learning defines pretraining tasks that can be learned without manual annotation, but require understanding of the data to be solved \cite{zhai2019s4l}. Recent research has shown that self-supervised learning is beneficial in datasets with limited labeled data for many different applications such as natural language processing \cite{devlin2018bert}, computer vision \cite{chen2020simple, he2020momentum}, robotics \cite{kahn2018self, sermanet2018time, deng2020self}, or protein modeling \cite{rao2019evaluating, rives2019biological}. Some research has tried to explain why self-supervised pretraining is particularly beneficial to train NNs. Using the lottery ticket hypothesis for self-supervised training in computer vision \cite{chen2020lottery} suggests that pretraining finds a more compact representation of the complex input data, which is beneficial for downstream tasks.

\section{Methodology}

\subsection{Problem Setting}

\noindent The goal of this work is to train an NN to recognize group activities during basketball games. In this section, we introduce notation and provide definitions used in the rest of the paper.

\noindent \textbf{Notation}

\noindent Since 2013, every NBA arena has a camera system to track the players (5 from each team) and the ball during basketball games. The system observes the 11 tracked objects by their locations in an x-y plane, where the x-axis goes across the length of the court and the y-axis goes from sideline to sideline. We pre-process the data, such that the offensive team attacks along the y-axis. We define the location of object $o$ at time step $t$  as  ${\boldsymbol \ell}_o^t = [x_o^t, y_o^t]$, with $o \in \mathcal{O} = \{B, A_1, \ldots, A_5, D_1, \ldots, D_5\}$ where $\mathcal{O}$ denotes the set of tracked objects, namely the ball $B$, each of the 5 attackers $\{A_1, \ldots, A_5\}$, and each of the 5 defenders $\{D_1, \ldots, D_5\}$. Using an ordered sequence of $L$ time frames, the trajectory of a tracked object $o$ can be expressed as ${\boldsymbol \tau}_o^{t-L+1:t} = [{\boldsymbol \ell}_o^{t-L+1}, \ldots, {\boldsymbol \ell}_o^{t}]$, with equally spaced time steps at an interval of $\Delta_t$.

\noindent \textbf{Trajectory Prediction Task}

\begin{figure}[t]
\begin{center}
\includegraphics[width=0.75\linewidth]{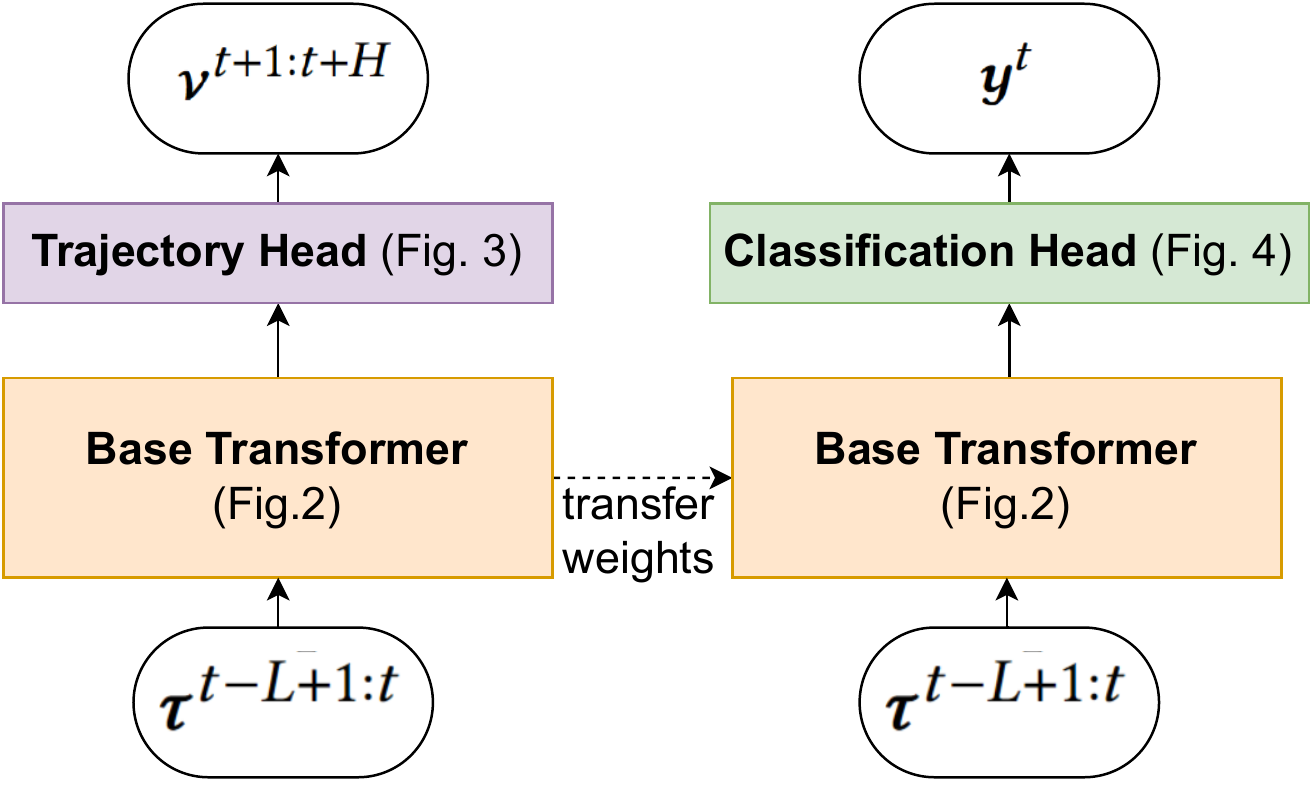}
\end{center}
\vspace{-2.5mm}
\caption{Illustration of the NETS architecture.}
\vspace{-5mm}
\label{fig:overview}
\end{figure}

\noindent We formulate the self-supervised task of predicting future trajectories for every tracked object $o$, represented by the vector ${\boldsymbol{\tau}}_o^{t+1:t+H} = [{\boldsymbol \ell}_o^{t+1}, \ldots,  {\boldsymbol \ell}_o^{t+H}]$, where $H$ is the number of future time steps (or prediction horizon) for which we predict the trajectory.

We denote as ${\boldsymbol \tau}^{t:t'}$ the list of trajectories for all objects, i.e. ${\boldsymbol \tau}^{t:t'} = [{\boldsymbol \tau}_{B}^{t:t'}, {\boldsymbol \tau}_{A_1}^{t:t'}, ..., {\boldsymbol \tau}_{D_5}^{t:t'}]$. Through a simple conversion, we can calculate velocity vector ${\boldsymbol \nu}^{t:t'}$ from the trajectory ${\boldsymbol \tau}^{t:t'}$. Although these two vectors are mathematically interchangeable, previous work on NBA data has shown that using velocity vectors as target to deep learning models is advantageous \cite{hauri2020multi}. This leads to the trajectory prediction task, where the objective is to predict ${\boldsymbol{\nu}}^{t+1:t+H}$ from ${\boldsymbol \tau}^{t-L+1:t}$. This is a self-supervised learning task, because future trajectories can be automatically extracted from the historical trajectory data and stored in dataset $\mathcal{D} = \{ ({\boldsymbol \tau}^{t-L+1:t}, {\boldsymbol{\nu}}^{t+1:t+H})| t = L, \ldots, T \text{-} H \}$, where $T$ is the total number of time steps.

\noindent \textbf{Group Activity Recognition (GAR)}

\noindent The objective of GAR is to predict group activity type $\boldsymbol{y}^t \in {1, ..., K}$ for each play sequence ${\boldsymbol \tau}^{t:t'}$, where K is the number of activity types. Similar to the trajectory prediction task, we can generate a labeled data set $\mathcal{D} = \{ ({\boldsymbol \tau}^{t-L+1:t}, {\boldsymbol{y}^t})| t = L, \ldots, T \}$, where each play sequence of $L$ time frames is matched to a label indicating the action type. The main difference compared to the trajectory prediction is that labels need to be provided externally.

\begin{figure}[t]
\begin{center}
\includegraphics[width=0.99\linewidth]{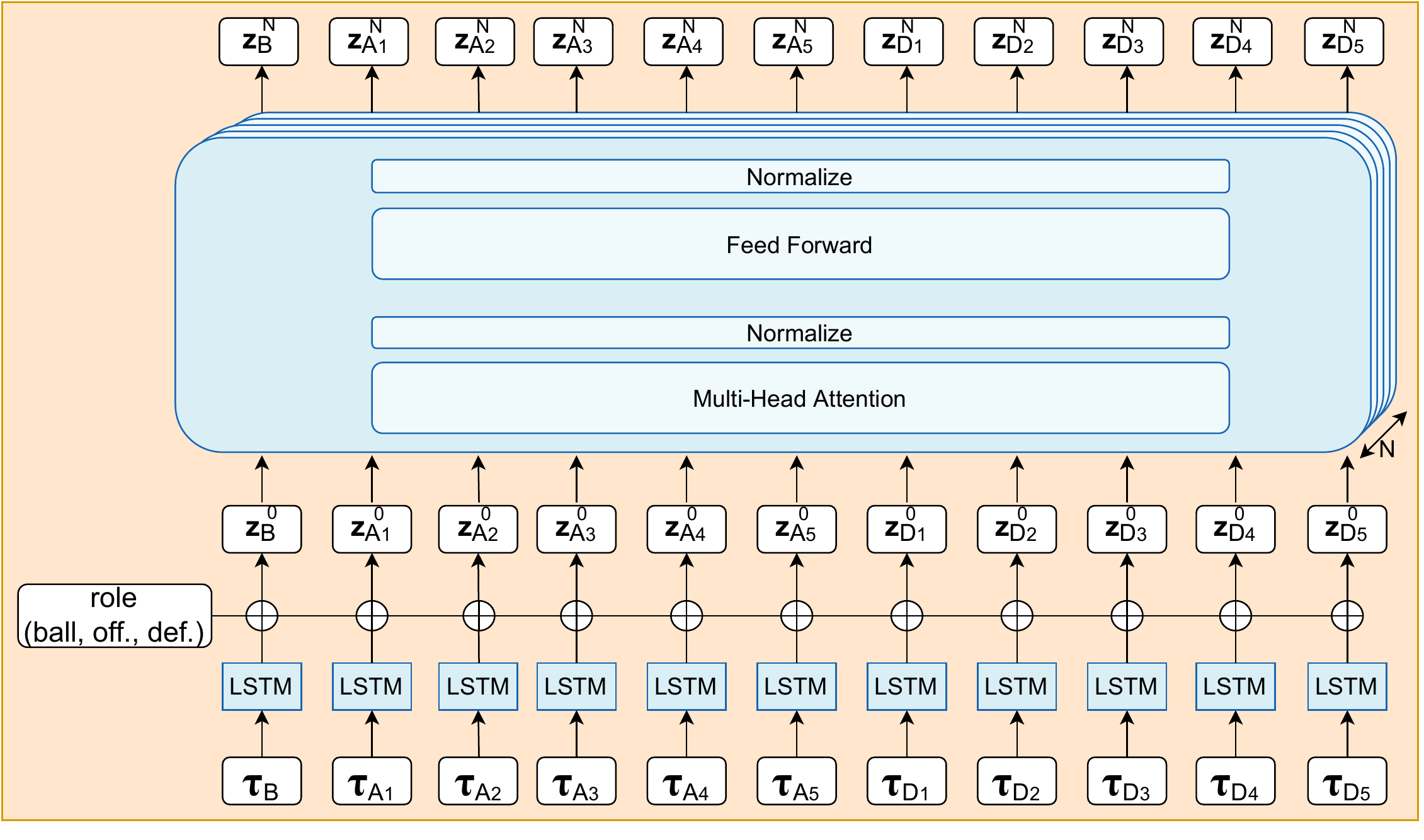}
\end{center}
\vspace{-4mm}
\caption{Base Transformer to generate embeddings. Trainable functions are in blue boxes. $\oplus$ stands for concatenation.}
\vspace{-1mm}
\label{fig:transformer_architecture}
\end{figure}

\begin{figure}[t]
\begin{center}
\includegraphics[width=0.9\linewidth]{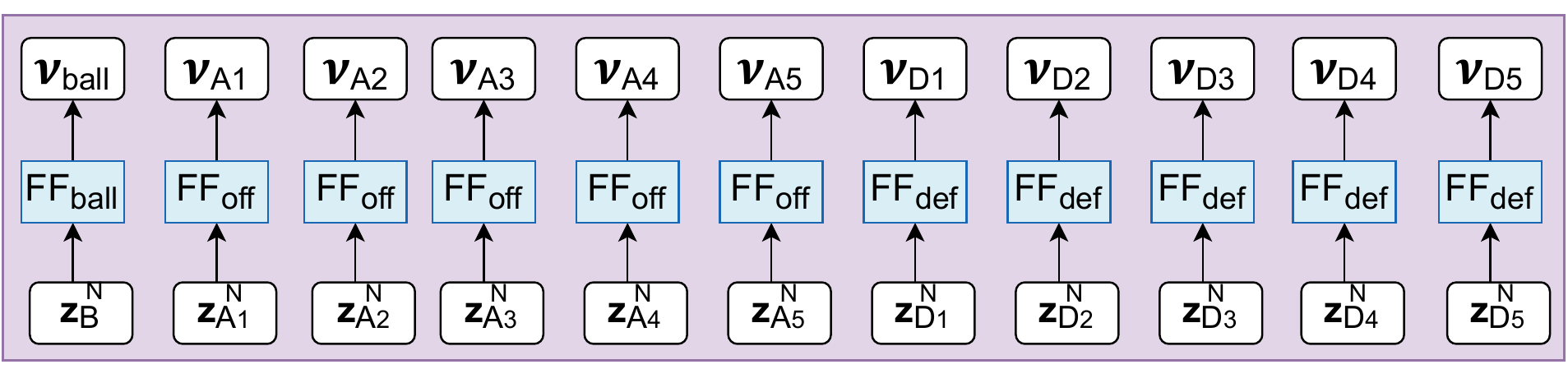}
\end{center}
\vspace{-4mm}
\caption{Trajectory head.}
\vspace{-1mm}
\label{fig:trajectory_architecture}
\end{figure}

\begin{figure}[t]
\begin{center}
\includegraphics[width=0.9\linewidth]{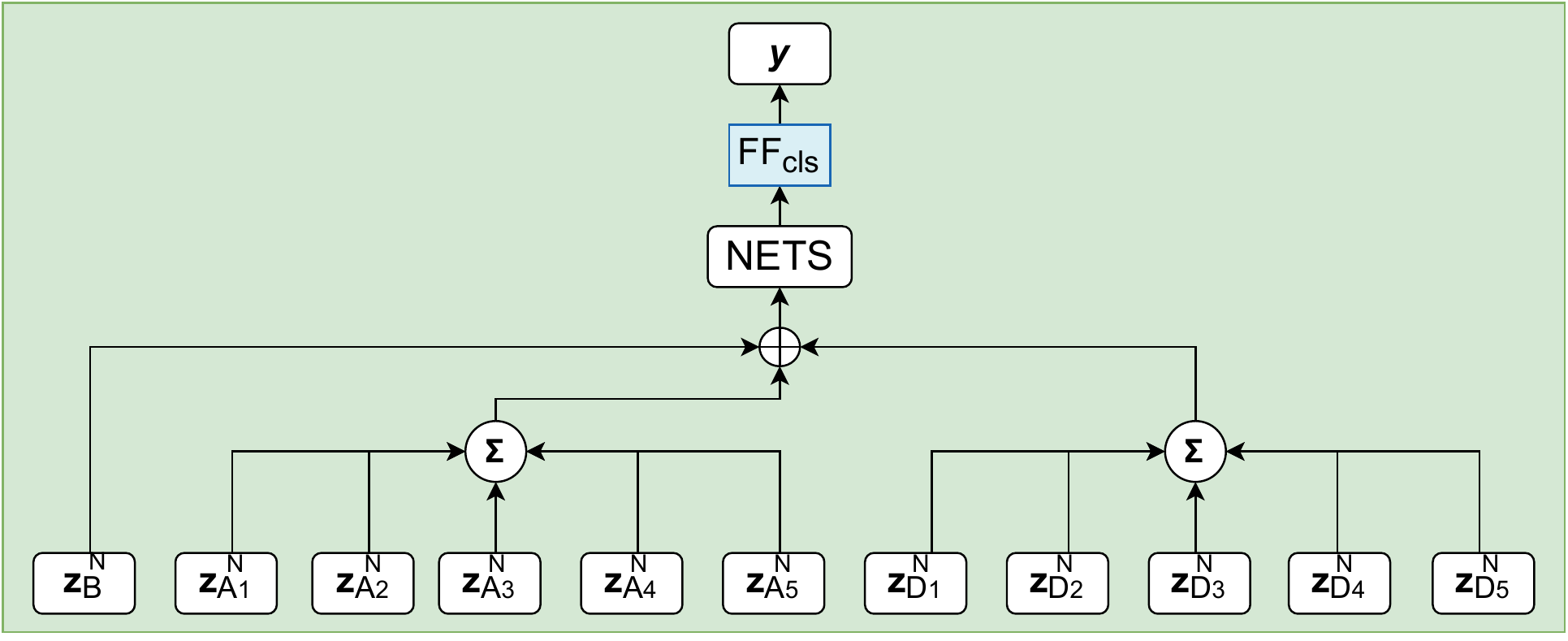}
\end{center}
\vspace{-4mm}
\caption{Classification head. $\sum$ stands for summation.}
\vspace{-2.5mm}
\label{fig:classification_architecture}
\end{figure}

\subsection{Framework}

\noindent A standard approach to train an NN for GAR, would be to use supervised learning on ground truth labels. However, as mentioned earlier, manual labeling is time consuming and expensive. We hypothesize that we can effectively train an NN through pretraining on the trajectory prediction task and fine-tune it using a large number of low-quality weak labels followed by fine-tuning using a small number of high-quality manual labels. To enable the pretrainaining and fine-tuning process, we utilize a modular NN architecture, which is described in Figure \ref{fig:overview}. For the trajectory prediction task, the trajectories ${\boldsymbol \tau}^{t-L+1:t}$ are input to the base model (we specify the chosen model in the  next section). The output of the transformer is then used as input to a trajectory prediction head for pretrainaining. The GAR task uses the same base model, but uses a classification head for activity classification. This modular approach allows the fine-tuning of weights after the pretrainaining of the trajectory prediction.

\subsection{Neural Network Architecture}
\label{NetworkArchitecture}

\noindent In this section, we will explain the detailed implementation of the base transformer, the trajectory head and the classification head.

\noindent \textbf{Transformer Encoder}

\noindent The input data consists of all 11 objects ${\boldsymbol \tau}_o^{t-L+1:t}$ of a given play sequence. Because these input vectors represent a time series for each tracked object $o$, we use a Long Short-Term Memory (LSTM) \cite{hochreiter1997long} layer to embed these trajectories into vectors (see Figure \ref{fig:transformer_architecture}), to fully exploit the temporal information of the input data.

The 11 objects have different properties since the ball behaves  differently than an attacker. We encode this role information as shown in Figure \ref{fig:transformer_architecture} and choose a one-hot positional encoding to differentiate each object class (i.e. the ball, offensive players, and defensive players). We concatenate these 3-dimensional vectors to the output of the LSTM layer, creating the input to the first attention layer $\boldsymbol{z}_o^0$. Although it would be possible to create embeddings for each player separately, previous work shows that adding player embeddings improves results only marginally while requiring a large feature engineering effort \cite{sicilia2019deephoops}.

To generate context-aware player embeddings, given the previously described input embeddings, we use a Transformer encoder with a multi-head self-attention mechanism \cite{vaswani2017attention}. Transformer encoders consist of multiple attention-based layers. Each layer learns to adjust the object representations in relation to other objects, where the objects in our application are the ball and the players. More formally, a Transformer processes the input $\boldsymbol{z}^l_o \in \mathbb{R}^{d_v}$ at layer $l$ to an output embedding $\boldsymbol{z}^{l+1}_o \in \mathbb{R}^{d_v}$, with input and output dimension $d_v$. The inputs are transformed into three matrices: query $Q$, key $K$, and value $V$, where each matrix represents the stacked input embeddings $\boldsymbol{z}^{l}_o$. These matrices are then transformed with trainable matrices $W_i^Q \in \mathbb{R}^{d_g \times d_v}$, $W_i^K \in \mathbb{R}^{d_g \times d_v}$, $W_i^V \in \mathbb{R}^{d_g \times d_v}$ and $W_i^O \in \mathbb{R}^{h d_v \times d_g}$, where $d_g$ is a model hyperparameter and $h$ is the number of self-attention heads. The multi-head self-attention function includes the residual connection and is calculated as

{\small
\vspace{-4mm}
\begin{align}
    \boldsymbol{z}^{l+1}_o &= LN(FF(LN(Att(Q, K, V)))) + \boldsymbol{z}^{l}_o \\
    Att(Q, K, V) &= Concat(head_1, ..., head_h) W^O \\
    head_i &= softmax \left( \frac{(Q W_i^Q) (K W_i^K)^T}{ \sqrt{d_k} } \right) V W_i^V ,
\end{align}
}
\noindent where $LN$ stands for layer normalization \cite{ba2016layer} and $FF$ is a fully connected feedforward network (see Figure \ref{fig:transformer_architecture}). We use the common practice of simplifying the hyperparameters by setting $d_g = d_v = d_k$, which we denote as hidden dimension $d_h$ from here on.

The transformer contains a stack of $N$ identical attention layers. Through this attention mechanism, the NN generates embeddings $\boldsymbol{z}^{N}_o$ for each tracked object $o$, taking into account the information of all other players and the ball.

\noindent \textbf{Trajectory Prediction Head}

\noindent After generating the embeddings for each tracked object, we can predict the future trajectories by using a FeedForward neural network (FF) to generate an output vector ${\boldsymbol \nu}^t_o$ for each tracked object (see Figure \ref{fig:trajectory_architecture}). Because the physical properties of motion for the ball are very different from humans and because the behavior of an offensive player is different from the behavior of a defensive player, we use separate FFs for each role to generate these trajectories. We denote FFs that share the same weights with the same subscripts in Figure \ref{fig:trajectory_architecture}.

\noindent \textbf{Classification Head}

\noindent To predict labels for GAR, we consider the context-aware transformer embeddings of the players and the ball from the last transformer layer. For GAR, the input order of the players within a team should not impact the assigned label and our model should be permutation invariant for each team. We use a team-level pooling layer which sums up all the embeddings of the players that belong to the offense and defense, respectively (see Figure \ref{fig:classification_architecture}). More formally, we the output values as

{\small
\vspace{-5mm}
\begin{equation}
    \boldsymbol{y} = softmax \left( FF \left(\boldsymbol{z}^N_B \oplus \sum_{i=1}^5 \boldsymbol{z}^N_{A_i} \oplus \sum_{i=1}^5 \boldsymbol{z}^N_{D_i}\right) \right),
\end{equation}
}

\noindent where $\oplus$ stands for concatenation. $\boldsymbol{y} \in \mathbb{R}^{K}$ is the output vector containing the probability for each of the $K$ predicted class.

We use a softmax activation function on the last layer so that the output can be interpreted as probabilities and trained by minimizing the Negative Log-Likelihood (NLL) loss for $K$ classes:

{\small
\vspace{-5mm}
\begin{equation}
    \mathcal{L}_\text{NLL} = -\frac{1}{ |\mathcal{D}| } \sum\limits_{\mathcal{D}} \sum\limits_{k=1}^K \alpha_k \boldsymbol{y}_k \ln(\boldsymbol{\hat{y}}_k),
\end{equation}
}

\noindent where $\alpha_k$ stands for class adjustments that balance the unequal class distribution \cite{king2001logistic}. $\boldsymbol{y}_k$ is the one-hot representation of ground truth, and $\boldsymbol{\hat{y}}_k$ is the predicted probability that the play belongs to class $k$.

\section{Experimental Setup}

\noindent In this section we describe the data set and the design of our labeling approach.

\subsection{Data Set}

\noindent We used publicly available movement data collected from 632 NBA games during the 2015-2016 season\footnote{https://github.com/sealneaward/nba-movement-data, last accessed August 2022; we are not associated with the data creator in any way.}. To avoid including play sequences where there is no significant action, we segmented the data into possessions, which start when the shot clock resets and end on the next reset. We only kept the part of the possession when all 10 players are in the offensive half of the court. Possessions shorter than $3s$ were discarded, resulting in 113,760 possessions. This amounts to 1.1 million seconds of game play where locations of the players and the ball are captured every $0.04s$. We downsampled the data by a factor of 3 to reduce computational cost (similar to \cite{zhan2018generating}) and obtain a sampling rate of $\Delta_t=0.12s$. For all models, we used $L=10$ time steps as input, corresponding to 1.2s of game play. We constructed a data set by splitting the possessions into 1.2s long non-overlapping segments, resulting in 869,905 play sequences. For each of those play sequences, we also included the future horizon of the next $H$ time steps, where we experimented with varying values of $H$ = 10, $H$ = 20, and $H$ = 40.

\subsection{Labeling Group Activities in Basketball}
\label{Labeling}

\noindent We generated labels for pick-and-rolls and handoffs, two commonly used tactics by NBA teams. Plays that do not fall into these two categories were labeled as "other", resulting in $K=3$ classes. Generating such labels manually is very time consuming because it not only requires watching entire basketball games, but also requires high concentration to follow multiple players at the same time. There are also many edge cases that require multiple viewings to decide on the correct label. We explain the weak-labeling process in detail in the Supplementary Material. In short, we used domain knowledge to generate programmatic rules to identify pick-and-roll and handoff behaviors.

\noindent \textbf{Weak-Labeling Pick-and-rolls}

\noindent Pick-and-roll is an offensive tactic in which the attacking team tries to block the defender guarding the ball handler. Another attacker (the so-called roll-man) helps the ball handler by standing in the way of the defender (see Figure \ref{fig:pick-and-roll}). This creates a difficult situation for the defender, who has to either run around the roll-man to keep guarding the ball handler or switch the assignment and guard the roll-man, which leads to a possible mismatch. In short, the rule-based approach matches each defender to an offensive player and then identifies situations in which the roll-man is very close to the ballhandler's defender (see details in Supplementary Material).

\noindent \textbf{Weak-Labeling Handoffs}

\noindent A handoff is a different offensive tactic in which two attackers cross paths, and the ball is handed off when the players are close to each other. The action can also involve a very short pass. This tactic allows one to give the ball to another player with a low risk of losing it, but it can also be used as an opportunity to stand in the defender's way. In this sense, it is closely related to a pick-and-roll, but the ball possession changes during the action. The rule-based approach to identifying a handoff tries to identify a change in possession between two players, with a small distance between the players and a short time of ball transition (see details in Supplementary Material).

The weak-labeling process produced 45,802 ``pick-and-rolls", 15,251 ``handoffs", and 808,852 ``other" play sequences.

\noindent \textbf{Manual Labeling}

\noindent We assigned manual labels for a total of 1,800 play segments. Because the data set is highly unbalanced according to weak-label distribution, we sampled play sequences equally from all 3 weak-labeled subgroups. To assign the labels, we generated video representations of 1.2s in length for each play segment. We carefully watched each video, replaying and freezing it until we were confident with its label. We sampled and labeled one video from each of the 3 weak labels until we found 600 pick-and-rolls manually labeled samples for each class. We used 50\% of these manual labels for testing and 50\% for fine-tuning our NETS model.

\section{Trajectory Prediction Problem}

\noindent We used the self-supervised trajectory prediction problem as a pretraining step for GAR. This task itself has found recent interest in the computer vision community \cite{sun2019stochastic, zhan2018generating, zheng2016generating, hauri2020multi}. In this section, we will show that our Transformer architecture can achieve state-of-the-art performance in trajectory prediction.

\subsection{Experimental Design}

\noindent We trained our models using pytorch. We used Adam optimizer with an initial learning rate of $5 \cdot 10^{-5}$ to reduce the mean squared error of predicted velocities

{\small
\vspace{-2.5mm}
\begin{equation}
\label{eq:MSE_loss}
\mathcal{L}_{\mathrm{MSE}}({\boldsymbol \nu}, \hat{\boldsymbol \nu}) = \frac{1}{2 H} \|{\boldsymbol \nu}-\hat{\boldsymbol \nu}\|_2^2 .
\end{equation}
}

We set all hidden dimension to $d_h=256$, and used a Transformer with $L=8$ layers and $h=64$ prediction heads (also see \ref{EvaluateWeakLabels}). We used a ReLU activation function between all layers except the output. We trained in batches of 512 samples. Early stopping was applied after 50 epochs if the validation error did not improve, which typically took around 400 epochs, corresponding to about 30 hours of training on a single GPU.

To evaluate the trajectory prediction task, different prediction horizons $H$ = 10, $H$ = 20 and $H$ = 40 were tested. From the predicted velocities, we computed the trajectories for the players and evaluated different approaches using two standard metrics \cite{gupta2018social, mozaffari2020deep} in trajectory prediction: average displacement error (ADE) and final displacement error (FDE). FDE is the expected euclidean distance between the predicted final location and the true final location of the tracked object after $H$ time steps. ADE describes the expected average euclidean distance between the predicted and the true trajectory at every predicted time step.

\subsection{Baselines}
\label{TrajectoryBaselines}

\noindent Since the trajectory prediction task has been studied in literature before, we compared our NETS architecture with several strong baselines.

\noindent \textbf{LSTM} The baseline LSTM refers to an approach using a 2-layer LSTM, with 128 hidden nodes in each layer as an input encoder, followed by a 2-layer FF network, also with 128 hidden nodes. We used the same loss function and training setup as for our NETS experiments.

\noindent \textbf{SocialGAN}\cite{gupta2018social} This model uses an LSTM-based generator, coupled with a social pooling layer, to account for nearby actors and a discriminator that learns to distinguish between actual and simulated trajectories. We used the same hyperparameter settings as in \cite{gupta2018social}. It is important to note that SocialGAN was designed for pedestrian prediction, where all objects are embedded the same way in a shared layer. Therefore, adding the ball location would require significant modification, which is why this model does not consider the ball's location in our experiments.

\noindent \textbf{M. VRNN} \cite{zhan2018generating} This model uses a hierarchical approach in which a Variatonal RNN is trained on programmatic weak supervision to first predict a location that a player wants to reach. Then it uses a second identical layer to predict a trajectory that the player will take to reach it. It is trained like a Variational Autoencoder, aiming to reproduce realistic behavior of offensive players. We used the same hyperparameter settings as in \cite{gupta2018social}. Because this baseline only predicts the trajectories of offensive players, we compared the performance of all models for the offensive players only.

\noindent \textbf{NETS$_{\text{no LSTM}}$} This is an ablation of our NETS model, with the only difference being that the input embedding is a 2-layer feedforward network with 256 nodes each, instead of the LSTM-layer in NETS. 

\subsection{Results}

\noindent We compared the performance of NETS on the trajectory prediction task to several strong baselines. Results for various prediction horizons $H$ are shown in Table \ref{tab:res_trajectory}. NETS$_{\text{no LSTM}}$ and NETS, which are based on transformers, outperformed the other baselines, showing improvements on different prediction horizons. The results held for both ADE and FDE, suggesting that the improvement can be observed along the entirety of a player's path. MACRO VRNNs trajectories are designed to resemble the behavior of basketball players, but can be far off from the ground truth. SocialGAN produced reasonably accurate predictions, but was inferior to other approaches that were specifically designed for this basketball dataset.

Another interesting finding is that NETS$_{\text{no LSTM}}$ performed worse than our full NETS model. We believe that the LSTM input embedding allows the model to more easily extract temporally dependent information in the input data, therefore explaining a reduced accuracy of a model that has to learn temporal patterns from unstructured input data.

\begin{table} [t]
\centering
\caption{Comparison of various models in terms of  error metrics ADE and FDE (in feet) for prediction horizons ${\boldsymbol H}$ = 10 (1.2 seconds), ${\boldsymbol H}$ = 20 (2.4 seconds), and ${\boldsymbol H}$ = 40 (4.8 seconds).}
\vspace{2.5mm}
\fontsize{8}{9}\selectfont
{
  \begin{tabular}{c rr rr rr}
     & \multicolumn{2}{c}{\bf ${\boldsymbol H}$ = 10}
     & \multicolumn{2}{c}{\bf ${\boldsymbol H}$ = 20}
     & \multicolumn{2}{c}{\bf ${\boldsymbol H}$ = 40}  \\
    {\bf Method} & {\bf ADE}  & {\bf FDE} & {\bf ADE}  & {\bf FDE} & {\bf ADE}  & {\bf FDE} \\
    \hline
    LSTM \cite{hauri2020multi} & 1.61 & 2.98 & 3.43 & 6.91 & 6.59 & 11.97 \\
    M. VRNN \cite{zhan2018generating} & 1.70 & 3.43 & 4.46 & 8.66 & 8.48 & 14.98 \\
    Social GAN \cite{gupta2018social} & 1.25 & 2.75 & 3.09 & 6.67 & 6.47 & 12.35 \\
    \hline
    NETS$_{\text{no LSTM}}$ & 1.18 & 2.51 & 2.98 & 6.42 & 6.31 & 11.84 \\
    NETS & \textbf{1.08} & \textbf{2.34} & \textbf{2.78} & \textbf{5.87} & \textbf{5.70} & \textbf{10.88} \\
    \hline
\end{tabular}
}
\vspace{-5mm}
\label{tab:res_trajectory}
\end{table}

\section{Group Activity Recognition (GAR)}

\noindent In this section, we report the performance of our proposed approach on GAR. We first evaluated the ability of these models to classify a large weak-labeled dataset. We were particularly interested to investigate if the self-supervised task of trajectory prediction can improve the accuracy of GAR. Then we evaluated our NETS model on manually labeled data to determine if a deep learning method can outperform the rule-based labeling approach.

\subsection{Experimental Design}

\noindent Due to our modular architecture design, the neural network architecture was the same as for the trajectory prediction task, except for the last prediction layer (see Section \ref{NetworkArchitecture}). We used an 80 / 10 / 10 train-, validation-, and test-split on all of these play sequences, and we applied early stopping after 50 epochs. To balance the distribution of 45,802 ``pick-and-rolls", 15,251 ``handoffs", and 808,852 ``other" play sequences, we downsampled other plays to 45,802 (same as pick-and-roll), we used weighting factors $\alpha_k$ of 0.77, 2.34, and 0.77 for pick-and-rolls, handoffs, and other, respectively.

We note that the group activity labels were dominated by play sequences labeled as 'other' and a trivial classifier predicting that the class would achieve 93.0\% test accuracy. Therefore, instead of reporting accuracy, we report the multi-class F1 score to compare the performance of the models. We calculated the confusion matrix $M$, where $M_{ij}$ represents the numbers of plays with ground truth labels $i$ classified as $j$. We calculated F1 scores by converting the 3-class classification problem into three binary classification problems, yielding the one vs. all F1-scores

{\small
\vspace{-5mm}
\begin{align}
    \text{precision}_i &= \frac{M_{ii}}{\sum_j M_{ji}}; \hspace{5mm} \text{recall}_i = \frac{M_{ii}}{\sum_j M_{ij}} \\
    \text{F}1_i &= 2 \cdot \frac{\text{precision}_i \cdot \text{recall}_i}{\text{precision}_i + \text{recall}_i}.
\end{align}
}

\subsection{Baselines}

\noindent We compared our NETS architecture to an \textbf{LSTM}-based neural network with the same settings described in Section \ref{TrajectoryBaselines}. To allow a fair comparison to our NETS architecture, we used the same training procedure of pretraining the model using the trajectory prediction task and then changing the prediction head to enable play classification.

Due to the lack of baselines using deep learning, we used standard shallow baselines, consisting of logistic regression \textbf{LReg}, random forest classifier \textbf{RForest}, and gradient boosting classifier \textbf{GBoost}. To establish a baseline, we trained 3 shallow models using the popular \textit{sklearn} implementation. The input to these shallow models were the play sequences ${\boldsymbol \tau}_o^{t-L+1:t}$ consisting of 220 features. We found the optimal parameter settings through grid search.

\subsection{Evaluation on Weak-Labels}
\label{EvaluateWeakLabels}

\noindent In Table \ref{tab:res_classification} we show F1 scores on the weak labeled data. Our NETS model outperformed shallow approaches by a large margin. The results show that the classification of group activities is a difficult problem for shallow models such as LReg, RForest, and GBoost, as indicated by the relatively low F1 scores of these traditional machine learning algorithms. Interestingly, we did not find that the LSTM baseline outperforms GBoost on the GAR problem.

\begin{table}[t]
\centering
\caption{Classification test performance compared to baselines. Tested on 4,581 pick-and-rolls (p\&r), 1,525 handoffs and 80,884 other plays.}
\fontsize{8}{9}\selectfont
{
  \begin{tabular}{l | rrr}
    & {\bf p\&r} & {\bf handoff} & {\bf other} \\
    \bf{Method} & \bf{F1-score} & \bf{F1-score} & \bf{F1-score}\\
    \hline
    LReg & 0.188 & 0.099 & 0.755 \\
    RForest & 0.329 & 0.261 & 0.810 \\
    GBoost & 0.398 & 0.443 & 0.915 \\
    LSTM & 0.360 & 0.490 & 0.905 \\
    NETS & \bf{0.856} & \bf{0.768} & \bf{0.988} \\
    \hline
\end{tabular}
}
\label{tab:res_classification}
\end{table}

\begin{table}[t]
\centering
\caption{Classification test performance of NETS with different architecture ablations. Tested on 4,581 pick-and-rolls (p\&r), 1,525 handoffs and 80,884 other plays.}
\fontsize{8}{9}\selectfont
{
  \begin{tabular}{ccc | rrr}
    & \bf{LSTM} &  & {\bf p\&r} & {\bf handoff} & {\bf other} \\
    \bf{pretrain} & \bf{embed.} & \bf{pooling} & \bf{F1-score} & \bf{F1-score} & \bf{F1-score}\\
    \hline
    \xmark & \xmark & \cmark & 0.705 & 0.667 & 0.973 \\
    \xmark & \cmark & \xmark & 0.777 & 0.644 & 0.980 \\
    \xmark & \cmark & \cmark & 0.802 & 0.675 & 0.982 \\
    \hline
    \cmark & \xmark & \cmark & 0.803 & 0.731 & 0.983 \\
    \cmark & \cmark & \xmark & 0.829 & 0.718 & 0.985 \\
    \cmark & \cmark & \cmark & \bf{0.856} & \bf{0.768} & \bf{0.988} \\
    \hline
\end{tabular}
}
\label{tab:res_classification2}
\end{table}

We performed an ablation study of the NETS models to evaluate our architecture design choices: We created a model ablation by removing the LSTM-embedding (see Figure \ref{fig:transformer_architecture}), in which the inputs were embedded with a 2-layer feedforward neural network. We also examined the influence of the team-wise pooling layer depicted in Figure \ref{fig:classification_architecture} by removing the summation and instead concatenating all representations $\textbf{z}_i^N$.

All the models in Table \ref{tab:res_classification2} were ablations of NETS and were based on a Transformer architecture. We see that all these models performed better than the shallow models or the LSTM-based model presented in Table \ref{tab:res_classification}. All models that were pretrained showed significant improvement compared to the same models without pretraining. As in the trajectory prediction task, we noted significant improvement when utilizing an LSTM-embedding at the input of the base Transformer instead of dense embedding layers. We also observed an improvement when using the team-pooling layer at the output. Furthermore, we observed that training GAR without pretraining took around 500 epochs when we trained from scratch but only roughly 200 epochs when we started with a pretrained model.

\subsection{Evaluation on Manual Labels}

\begin{table}[t]
\centering
\caption{Confusion matrix of weak-labels tested on manually labeled data, consisting of 300 pick-and-rolls (p\&r), 300 handoffs and 300 other plays.}
\fontsize{8}{9}\selectfont
{
  \begin{tabular}{l | rrr}
   & \multicolumn{3}{c}{\bf ground truth} \\
    & {\bf p\&r} & {\bf handoff} & {\bf other} \\
    \hline
    {\bf predicted p\&r}  & 282 & 43 & 24 \\
    {\bf predicted handoff} & 5 & 253 & 21 \\
    {\bf predicted other} & 13 & 4 & 260 \\
    \hline
\end{tabular}
}
\vspace{2mm}
\label{tab:confusion_table}
\end{table}

\begin{table}[t]
\centering
\caption{Accuracies of four variants of NETS. "weak-labels" refers to concordance between weak- and manual labels. Tested on manually labeled data, consisting of 300 pick-and-rolls (p\&r), 300 handoffs and 300 other plays.}
\fontsize{8}{9}\selectfont
{
  \begin{tabular}{l | rrr}
    & {\bf p\&r} & {\bf handoff} & {\bf other} \\
    \bf{method} & \bf{F1-score} & \bf{F1-score} & \bf{F1-score}\\
    \hline
    1) weak-labels & 0.869 & 0.874 & 0.893 \\
    2) NETS & 0.915 & 0.863 & 0.908 \\
    3) NETS, finetune on ml$^*$ only & 0.784 & 0.813 & 0.844 \\
    4) NETS, validate on ml$^*$ & 0.932 & 0.930 & 0.908 \\
    5) NETS, finetune on ml$^*$ & 0.951 & 0.938 & 0.902 \\
    \hline
    \multicolumn{4}{l}{$^*$ ml = manual labels}
\end{tabular}
}
\label{tab:res_manual}
\end{table}

\begin{figure*}[t]
\centering
\subfigure[Pretraining on trajectory prediction]{
\label{fig:tsne_a}\includegraphics[width=0.27\linewidth]{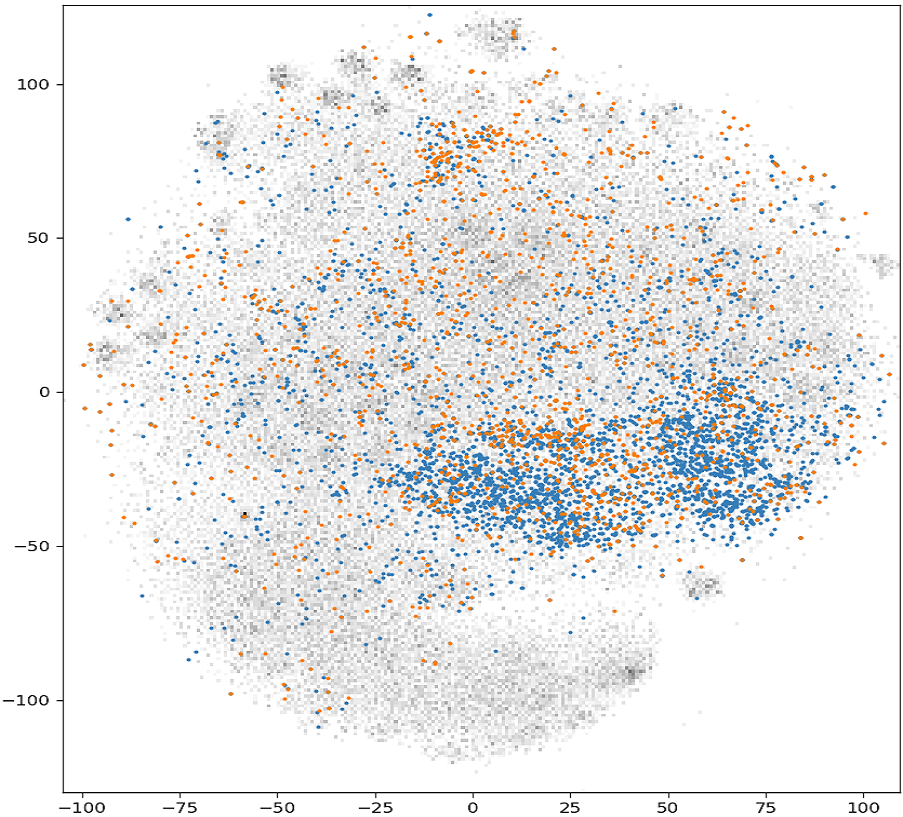}
}\hfill
\subfigure[Pretrained on trajectory prediction and fine-tuned on pick-and-roll classification.]{
\label{fig:tsne_c}\includegraphics[width=0.27\linewidth]{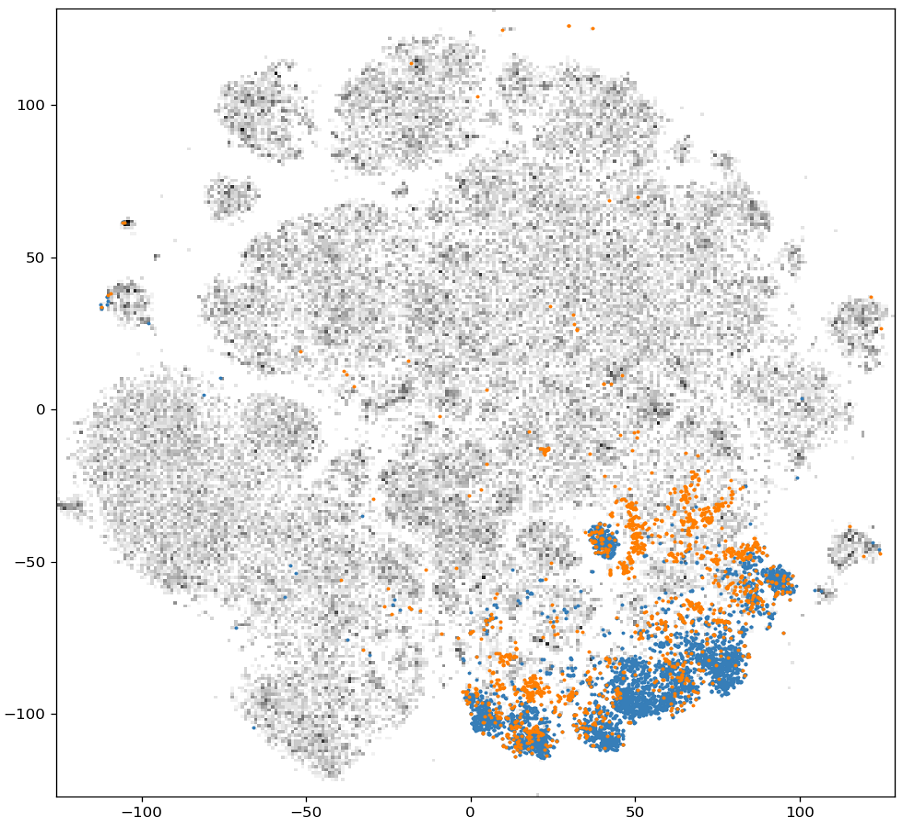}
}\hfill
\subfigure[Pretrained on trajectory prediction, and fine-tuned on pick-and-roll and handoff classification.]{
\label{fig:tsne_d}\includegraphics[width=0.27\linewidth]{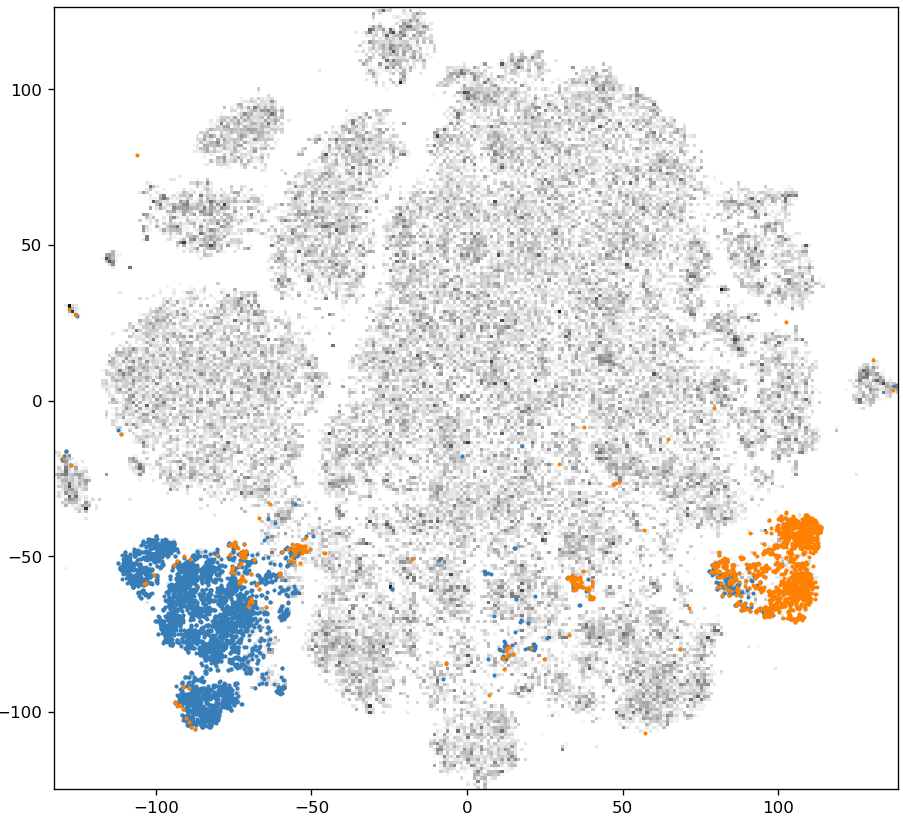}
}
\subfigure{\raisebox{30mm}{
\includegraphics[width=0.08\linewidth]{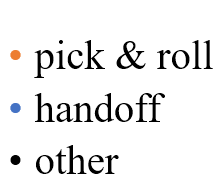}
}
}
\caption{t-SNE of embeddings of - pick-and-rolls (blue), - handoffs (orange) and - random plays (black).}
\label{fig:tsne}
\end{figure*}

\begin{figure}[t]
\centering
\includegraphics[width=0.6\linewidth]{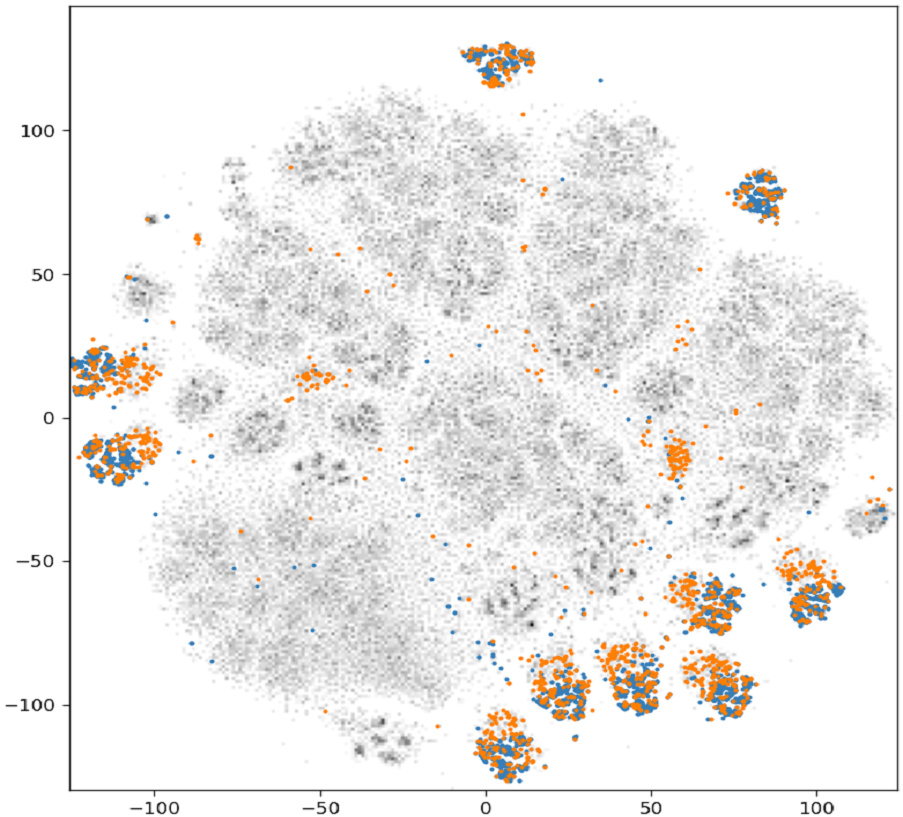}
\caption{t-SNE of embeddings of - pick-and-rolls (blue), - handoffs (orange) and - random plays (black). Pretrained on trajectory prediction and fine-tuned on pick-and-roll classification without team-wise pooling.}
\label{fig:tsne_b}
\end{figure}

\noindent As described in Section \ref{Labeling}, we manually labeled 600 pick-and-rolls, 600 handoffs and 600 other plays, which we split into 50\% train- and 50\% test-set.

To get an insight into the performance of our rule-based weak-labeling approach, we show the confusion table of the weak-labels compared to the assigned ground truth from manual labeling in Table \ref{tab:confusion_table}. While the weak-labeling rules tended to extract many correctly labeled play sequences, both rules for pick-and-rolls and handoff included 24 and 21 play sequences that were labeled as other by manual labeling. Furthermore, there were relatively frequent misclassification of handoffs as pick-and-rolls.

In the previous subsection, we showed that our NETS architecture can predict the weak-labels with high accuracy. Next, we evaluated the performance on manual labels. We also examined the usefulness of weak and manual labels when testing on these manual labels.

We evaluated the following approaches: 1) Using the rule-based weak-labeling approach to classify the group activities, which requires human expertise to define the rules and is not a deep learning approach. 2) Pretrain the NETS model on a trajectory prediction task and finetune the model using a large amount of weak-labels. 3) Pretrain the NETS model on a trajectory prediction task and finetune the model only using 900 manual labels. 4) Pretrain the NETS model on a trajectory prediction task, use the weak-labels in the training set and the manual labels in the validation set for early stopping. 5) Pretrain the NETS model on a trajectory prediction task, finetune it first on the weak-labels and then finetune again on the manual labels.

The results in Table \ref{tab:res_manual} show the F1 scores for the rule-based weak-labels calculated from the confusion matrix (Table \ref{tab:confusion_table}). We observed a slight improvement when training NETS on the weak labels compared to the weak labels themselves, which we hypothesize stems from the ability of the neural network to generalize better than the rules alone. In contrast, finetuning only on manual labels performed worse than the rule-based labeling, indicating that we did not have enough manual labels to train a large NN. Using manual labels in the validations set further improved the accuracy, implying that NETS overfit the weak labels without access to manual labels. Using a sequential finetuning paradigm resulted in the highest accuracy (row 5). When comparing row 5) with the weak-label F1 scores in row 1), we can observe that the accuracy on pick-and-rolls increased from 0.869 to 0.951, for handoffs it increased from 0.874 to 0.938, and for others from 0.893 to 0.902.

\section{Evaluation of Representations}
\label{EvaluateRepresentations}

\noindent In this section, our goal is to show that the learned embeddings got increasingly better with each learning step.

For qualitative analysis, we transformed the higher dimensional embeddings into 2D space and used t-SNE \cite{van2008visualizing} on the test set. We then visually examined how tightly pick-and-rolls and handoffs were packed together. Scatterplots shown in Figure \ref{fig:tsne} illustrate the difference between internal representations of the three NETS models.

Figure \ref{fig:tsne_a} shows the embeddings generated from a model trained on the trajectory prediction task. The embeddings of pick-and-roll plays and handoffs did not seem to be clustered together and were scattered within other plays. Some clusters were forming, but they were not very distinct.

Figure \ref{fig:tsne_c} shows that pick-and-rolls are clearly separated from other plays. More interestingly, handoffs are often clustered together and are mostly placed between pick-and-rolls and other plays, correctly indicating that there are some similarities between the plays. It also shows some unwanted behavior since many handoffs are clearly placed among pick-and-rolls, indicating that it is difficult to distinguish the two types of play from each other.

Figure \ref{fig:tsne_d} shows the embeddings created after fine-tuning on both pick-and-rolls and handoffs. The two types of strategies were now clearly separated, with only few exceptions. Furthermore, we can see more distinct clusters forming among the other plays, indicating that the embeddings were of higher quality and more capable of distinguishing different types of strategies, even for those that have no labels.

\textbf{Impact of Pooling Layer}

Figure \ref{fig:tsne_b} shows the clustering after training the model both on the trajectory prediction task and pick-and-roll classification (but without training on handoffs). The model in Figure \ref{fig:tsne_b} did not use team-wise pooling (Figure \ref{fig:tsne_c} shows the corresponding plot with team-wise pooling). We observe that pick-and-rolls are separated from random plays in both pictures. However, without team-wise pooling we observe the formation of 10 distinct clusters with sub-clusters. Through further analysis, we found that the clusters correspond to the player indexes of the 2 involved offensive players (in a basketball team, there are ${5 \choose 2} = 10$ combinations of 2 players). On one hand, this indicates that the model successfully learned to identify the involved players, but on the other hand it uses a lot of resources to indicate at which input position that player is.

\section{Conclusion and Future Work}

\noindent We have shown that our NETS model with a specifically designed transformer architecture to address common challenges in sports analytics can classify group activities with high accuracy. Pretraining the model on a self-supervised trajectory prediction task significantly improved the model performance on downstream tasks. In future work, we aim to utilize the generated play embeddings to discover strategic differences employed by different teams and generate a tool that can help automatically analyze a game.

{\small
\bibliographystyle{ieee_fullname}
\bibliography{_main.bib}
}

\clearpage
\appendix

\section{Labeling Process}

A general approach to creating a labeled data for different types of plays is to do it manually. However, it is very time consuming and tedious. Instead, we develop an algorithm for automatic labeling, which results in a large number of labels. For any complex strategy, there is a need to observe multiple frames in order to assign a label.

A reasonable strategy for automatic labeling is to find carefully crafted rules with reasonable coverage and precision. Manual verification over a small number of labels can provide sufficient feedback for the quality of a rule.

We first find rules to detect basic concepts, such as ball possession and defensive assignments. We say that a player $p_i$ has possession of the ball if he is the closest player to the ball for at least 5 consecutive frames, the ball is within 5 ft of the player, the ball is lower than 10 ft from the ground, and the ball's speed is slower than 25 ft/s. Next, we find defensive assignments by matching each defender with an offensive player. This is a linear sum assignment problem, where the defense is trying to reduce the total sum of euclidean distances between each defender and his assigned offensive player.

\textbf{Pick-and-rolls}

A pick-and-roll is an offensive tactic in which the attacking team tries to block the defender guarding the ball handler. Another attacker (the so called roll-man) helps the ball handler by standing in the way of the defender. This creates a difficult situation for the defender, who has to either run around the roll-man to keep guarding the ball handler or switch the assignment and guard the roll-man, which leads to a possible mismatch.

We denote the ball handler as player $a_1$ and set up rules to detect the moment when the roll-man $a_2$ blocks the path of the ball handler’s defender $d_1$. The three players form a triangle with side lengths $\delta_a$ between the two offensive team mates, $\delta_{d1}$ between the defender and the ball handler, and $\delta_{d2}$ between the defender and the roll-man. During a pick-and-roll, all these three players should be close together, so we set a pick-and-roll label when these values go below a certain threshold. We validated different thresholds by manually checking all generated pick-and-rolls during an entire game and found that the best results are when the thresholds for $\delta_a$ and $\delta_{d1}$ are set to 6 ft and the threshold for $\delta_{d2}$ is set to 3 ft. Although the rule involving these distances is based on a single frame, the rule that decides whether a player is a ball handler involves a possession of at least 5 frames.

A manual inspection of 200 plays confirmed that 164 plays were actually pick-and-rolls, giving a specificity of 82\%. Some observed errors are situations where two attackers are close to each other without performing a pick-and-roll. Although rare, this can happen in a handoff (see below), since it is possible for a handoff and pick-and-roll to happen within a few frames of each other.

\textbf{Handoffs}

A handoff is a different offensive tactic in which two attackers cross paths and the ball is handed off when the players are very close to each other. The action can also involve a very short pass. This tactic is used to give the ball to the best player with a low risk of losing the ball (compared to a longer pass), but can also be used as an opportunity to set a screen (i.e. stand in the way of the defender) by the initial ball handler. In this sense, it is closely related to a pick-and-roll, but the ball possession changes during the action.

To detect a handoff, we try to find the frame when the ball changes possession between two offensive players. We observe that a handoff can be executed simply by stretching out the ball to another player, and that the average wingspan of an NBA player is rather large-- about 6'6''. Therefore we define a key frame as a handoff if the possession has changed between two offensive players and the two players are closer than 6.5 ft apart. Manual evaluation of 200 labeled handoffs show 181 real handoffs for a specificity of 90.5\%.

\section{Ablation on Hyperparameters}

\begin{table}[h]
\centering
\caption{Classification test performance for models with different parameter settings, pretrained on trajectory  with prediction horizon $H$.}
\fontsize{8}{9}\selectfont
{
  \begin{tabular}{cccc | rrr}
    {\bf $N$} &  & {\bf $h$} &  & {\bf p\&r} & {\bf handoffs} & {\bf other} \\
    {\bf layers}  & {\bf $d_h$} & {\bf heads} & {\bf $H$} & \bf{F1-score} & \bf{F1-score} & \bf{F1-score}\\
    \hline
    16 & 256 & 32 & 20 & 0.841 & 0.758 & 0.986 \\
    16 & 256 & 128 & 20 & 0.846 & 0.762 & 0.985 \\
    8 & 256 & 64 & 20 & 0.849 & \bf{0.769} & 0.987 \\
    \hline
    16 & 256 & 64 & 10 & 0.838 & 0.743 & 0.983 \\
    16 & 256 & 64 & 20 & \bf{0.856} & \bf{0.768} & \bf{0.988} \\
    16 & 256 & 64 & 40 & 0.831 & 0.719 & 0.981 \\
    \hline
\end{tabular}
}
\label{tab:res_ablation}
\end{table}

Table \ref{tab:res_ablation} shows classification results on 3-way classification problem that was used to finetune the hyperparameters. To simplify the hyperparameter tuning process, we set all hidden dimensions $d_g$, $d_k$ and $d_v$ in the attention layers, as well as the hidden dimensions of LSTM and fully connected layers to the same value $d_h = 256$. All of the models were pretrained on the trajectory prediction task and then fine-tuned, with the only difference between the models being their hyperparameters. We find that hyperparameter tuning generally does not have a large impact on the resulting performance, but that using larger values for the number of layers $N$ and the hidden dimension $d_h$ improve the results. The best results are achieved with $h = 64$ attention heads and a prediction horizon $H$ = 20.

\end{document}